\documentclass[11pt]{article}

\usepackage[preprint]{acl}

\usepackage{times}
\usepackage{latexsym}
\usepackage{marvosym}

\usepackage[T1]{fontenc}

\usepackage[utf8]{inputenc}

\usepackage{microtype}

\usepackage{inconsolata}

\usepackage{graphicx}
\usepackage{amsmath}
\usepackage{float}
\usepackage{booktabs} 
\usepackage{multirow}
\usepackage{algorithm}
\usepackage{algpseudocode}
\usepackage{comment}
\usepackage{xcolor}
\newcommand{\best}[1]{\textbf{#1}}
\newcommand{\second}[1]{\underline{#1}}
\title{Test-Time Latent Computation to Improve Visual Perception in Multimodal Large Language Models}

\author{
  Chen Yang\textsuperscript{1},
  Zheng Fang\textsuperscript{1},
  Wenqi Fan\textsuperscript{2},
  Jun Xia\textsuperscript{1,\Letter} \\
  {\textsuperscript{1}The Hong Kong University of Science and Technology (Guangzhou)} \\
  {\textsuperscript{2}The Hong Kong Polytechnic University} \\
  {\small
    \href{mailto:cyang931@connect.hkust-gz.edu.cn}{cyang931@connect.hkust-gz.edu.cn};
    \href{mailto:zfang723@connect.hkust-gz.edu.cn}{zfang723@connect.hkust-gz.edu.cn}} \\
  {\small
    \href{mailto:wenqifan03@gmail.com}{wenqifan03@gmail.com};
    \href{mailto:junxia@hkust-gz.edu.cn}{junxia@hkust-gz.edu.cn}}
}

\begin{document}
\maketitle
{\renewcommand{\thefootnote}{\Letter}%
\footnotetext[1]{Corresponding author: Jun Xia.}}
\begin{abstract}
Recent work on Large Language Models (LLMs) suggests that reasoning capabilities can be enhanced at test time not only through longer token-level reasoning traces, but also through additional computation in latent space. This raises a natural question for Multimodal Large Language Models (MLLMs): can additional computation in latent space improve visual perception at test time? In this work, we first verify that MLLMs contain stable reusable layers that can be re-forwarded at inference time without disrupting normal generation. Based on this observation, we propose \textbf{Lo}ok \textbf{Ba}ck in the \textbf{L}atent \textbf{S}pace (\textbf{Lobals}), a training-free decoding method that identifies the contiguous layer window with the strongest visual-token attention and re-forwards through it, allowing the model to revisit visual evidence in latent space. We demonstrate that this latent-space revisit strategy can improve MLLMs' performance across diverse multimodal settings, including general vision-language tasks, visual hallucination tasks, and vision-centric tasks. Code is available \href{https://github.com/wuycM/Lobals}{here}.
\end{abstract}

\section{Introduction}

Transformer language models are typically executed as a fixed sequence of layers, with each layer applied once during inference~\cite{devlin2019bert,brown2020language,liu2024llavanext}. Despite sharing the same architectural design, layers in Large Language Models (LLMs) exhibit structured functional differences across depth~\cite{yu2026discovering,csordas2026language,sun2024transformer,geva2023dissecting,yang2025internal}. Probing studies show that linguistic information emerges progressively, with lower layers capturing surface features, middle layers encoding syntactic and semantic abstractions, and higher layers becoming more tied to output prediction~\citep{tenney2019bert,jawahar2019does}. Lens-based analyses further suggest that predictions are gradually refined along the layer stack, indicating that LLMs follow implicit computational trajectories during inference~\citep{nostalgebraist2020logitlens,belrose2023eliciting}. Moreover, recent studies find that middle layers often form a shared semantic space, supporting their role as a central region for reasoning~\citep{wu2025semantic, wendler2024llamas, ng2026sapirwhorf}.

This layer-wise structure has motivated recent efforts to increase computation in latent space instead of generating longer token-level reasoning traces. 
Training-based methods have explored recurrent and looped computation, showing that additional internal computation can strengthen reasoning by refining hidden representations before output generation~\citep{zhu2025scaling,geiping2026scaling,chen2025inner}. While these approaches typically require specialized architectures or additional training, some works~\cite{ng2026rys,li2025skiplayerloopit} show that a frozen LLM can benefit from simply repeating a contiguous block of middle layers at inference time. This suggests that middle layers may function as reusable latent thinking regions, and that revisiting selected intermediate layers can provide an effective mechanism for strengthening reasoning without parameter updates.
\begin{figure*}[t]
    \centering
    \includegraphics[width=\textwidth]{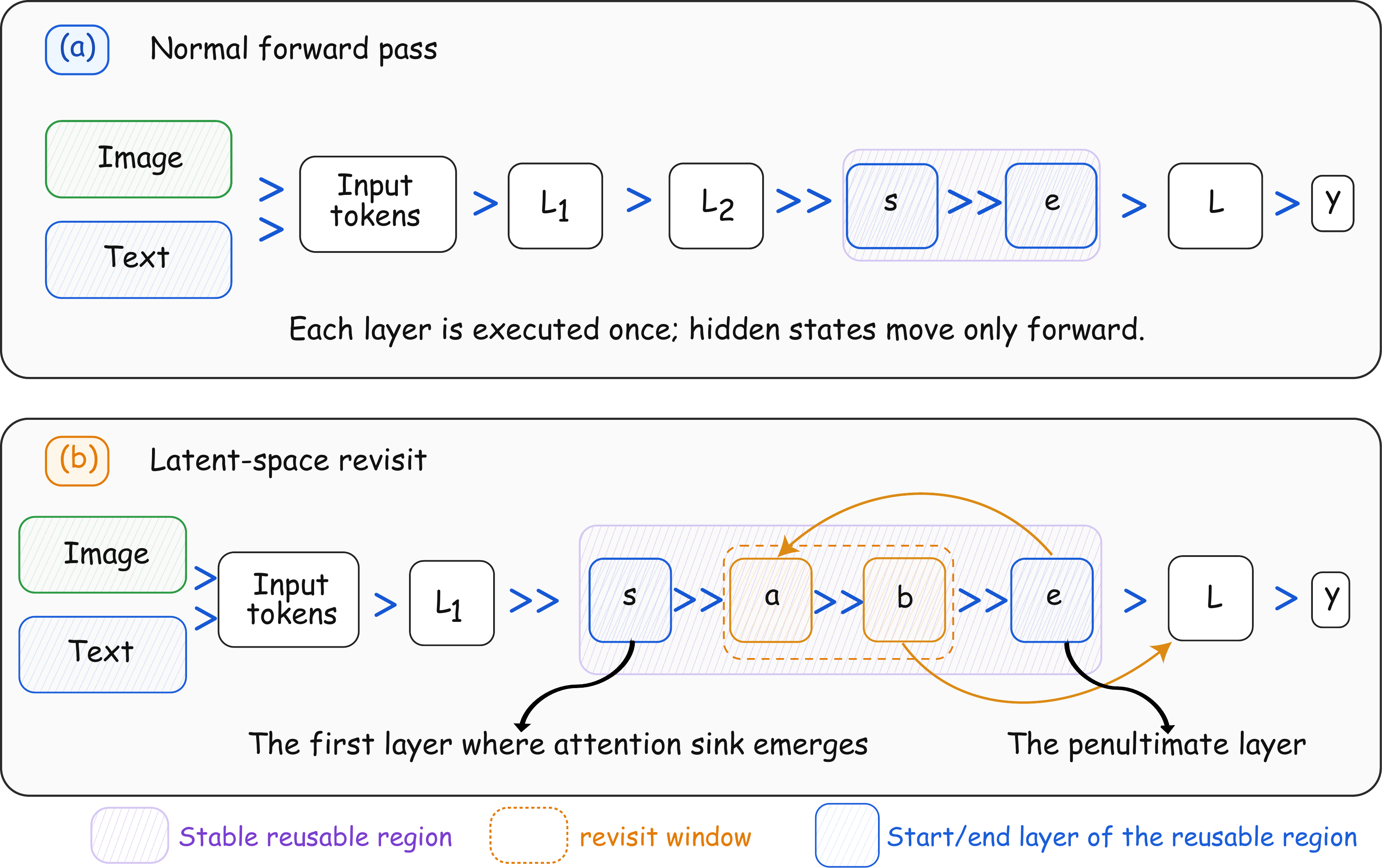}
    \caption{
    Illustration of latent-space revisit. 
    (a) In a normal forward pass, hidden states are propagated layer by layer, and each transformer layer is executed only once. 
    (b) In latent-space revisit, the hidden state from the penultimate layer is fed back into an earlier window within the stable reusable region.
    }
    \label{fig:overview}
\end{figure*}

Current mainstream Multimodal Large
Language Models (MLLMs) follow the LLaVA-style architecture~\cite{liu2023visual}, inheriting the Transformer backbone of LLMs while introducing visual tokens through a visual encoder and projection module~\cite{bai2025qwen3, chen2024internvl}. 
Recent studies suggest that, similar to LLMs, MLLMs also exhibit structured layer-wise behavior: shallow layers aggregate global visual information into token embeddings, middle layers are the primary region for cross-modal interactions between visual evidence and the query, and deep layers refine the integrated multimodal representation into output-oriented features~\citep{yin2025clearsight,zhang2025cross,kaduri2025s}. 

This raises a natural question: if the middle layers of LLMs serve as key regions for thinking and can be revisited to improve reasoning, can MLLMs similarly enhance visual perception by re-forwarding through their visual-textual interaction regions? The answer is yes. As a first step, we verify that, similar to LLMs, MLLMs also contain stable reusable regions that can be re-forwarded at inference time while preserving normal generation. These regions are mainly concentrated in the middle-to-late layers, as shown in Figure~\ref{fig:overview}. More importantly, we observe a consistent pattern across different MLLMs: the reusable region starts from the first layer where the attention-sink pattern~\citep{ICLR2024_5e5fd18f} emerges and ends at the penultimate layer. For instance, in Qwen3-VL-8B~\cite{bai2025qwen3}, which contains 36 transformer decoder layers, the stable reusable range spans layers 8 to 35.

Based on this observation, we propose \textbf{Lo}ok \textbf{Ba}ck in the \textbf{L}atent \textbf{S}pace (\textbf{Lobals}), a training-free decoding method for MLLMs. During generation, Lobals guides the model to revisit the latent region that most strongly attends to visual evidence. Specifically, we measure the attention assigned to visual tokens across reusable layers at each decoding step, identify the contiguous layer window with the highest visual-token attention ratio, and re-forward through this window. This lets the model look back at the image in latent space, strengthening visual perception without modifying model parameters or introducing extra tools.

Our contributions are summarized as follows:
\begin{itemize}
    \item We identify stable reusable regions in MLLMs where revisiting computation preserves normal generation. Across models, these regions consistently span from the first layer exhibiting the attention-sink pattern to the penultimate layer. We further provide empirical evidence for why they can be safely reused.
    \item We propose Lobals, a training-free decoding method that selects the contiguous layer window with the strongest visual-token attention ratio and re-forwards through it to revisit visual evidence in latent space.
    \item Extensive experiments suggest that \textbf{Lobals} can improve MLLMs' performance across diverse multimodal settings, including general vision-language tasks, visual hallucination tasks, and vision-centric tasks, without modifying model parameters or introducing extra tools.
\end{itemize}

\section{Related Work}

\textbf{Interpretability of LLMs and MLLMs.} Recent interpretability studies suggest that transformer computation is not uniformly distributed across depth. In LLMs, probing and lens-based analyses show that linguistic features and token predictions emerge progressively, with middle layers often serving as a semantic bottleneck or latent reasoning region~\citep{wu2025semantic,nostalgebraist2020logitlens,wendler2024llamas}. In MLLMs, recent work extends this layer-wise view to cross-modal computation. Analyses of cross-modal information flow show that visual evidence is integrated unevenly across depth, with the middle-to-late layers playing a particularly important role in multimodal interaction~\citep{neo2025towards,yin2025lifting,zhang2025cross,kaduri2025s,wang2025mllm}.

\textbf{Recurrent Computation in Language Models.}
Recurrent computation in language models has evolved from memory recurrence for extending context~\cite{dai-etal-2019-transformer, NEURIPS2022_47e28862} to depth-wise recurrence and parameter sharing that reuse Transformer computation for iterative refinement~\cite{DBLP:journals/corr/abs-1807-03819, ng2024loopneuralnetworksparameter}. Recent works further connect recurrence with test-time compute scaling in LLMs. One line of work trains models to perform additional computation in latent space by recurrently applying shared or selected Transformer blocks, thereby increasing effective depth without proportional parameter growth~\cite{geiping2026scaling,chen2025inner,zhu2025scaling}. Complementary to these training-based methods, training-free approaches show that pretrained LLMs already contain reusable layer structures, where inference-time computation paths can be adapted by revisiting intermediate layers without modifying model weights~\cite{li2025skiplayerloopit, ng2026rys}. Compared with recurrent computation in LLMs, recurrent computation in MLLMs remains underexplored. Lobals fills this gap by enabling MLLMs to revisit visually grounded latent windows to improve visual perception in a training-free manner.

\section{Observation}

We first investigate whether MLLMs contain latent regions that can be revisited without disrupting normal generation. Specifically, we conduct this analysis on Qwen3-VL-2B/8B/32B~\cite{bai2025qwen3} and InternVL3.5-14B~\cite{wang2025internvl3} to cover models with different scales and architectures. For each model, we sample 500 images from COCO~\cite{lin2014microsoft} and use the prompt "Describe the image." During the normal forward pass, we obtain the hidden state after layer $n$. We then feed this hidden state back into each possible earlier starting layer $m \in \{1,\dots,n\}$, pass it again through the consecutive layers $[m,n]$, and resume the standard forward computation from layer $n+1$ onward. The generated responses are evaluated by GPT-5.1 for fluency. If any sampled image leads to a non-fluent output, such as repetitive text or garbled characters, the corresponding revisit range is considered to disrupt normal generation.

\begin{figure}[h]
    \centering
    \includegraphics[width=\columnwidth]{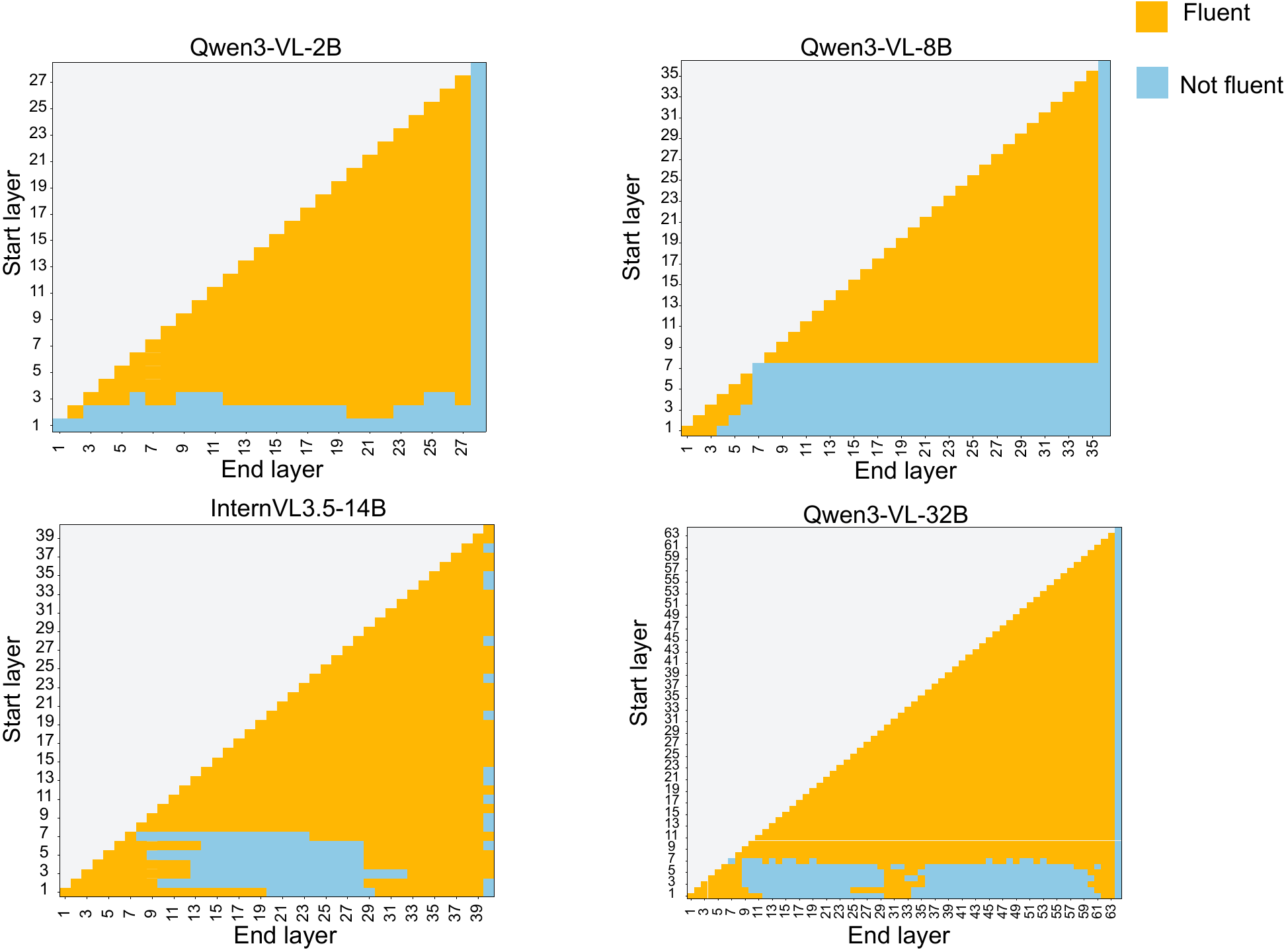}
    \caption{
Fluency analysis of latent-space revisit across different MLLMs. 
Each cell represents a revisit range, where the hidden state after an endpoint layer is fed back into an earlier starting layer and the model then resumes normal decoding. Yellow indicates that all sampled generations are fluent, while blue indicates that at least one generation is non-fluent. 
}
    \label{fig:fluent}
\end{figure}
The results are shown in Figure~\ref{fig:fluent}. We observe that each model contains its own stable reusable region. Specifically, the reusable range is $[4,27]$ for Qwen3-VL-2B, $[8,35]$ for Qwen3-VL-8B, $[8,63]$ for Qwen3-VL-32B, and $[8,39]$ for InternVL3.5-14B. In contrast, revisit ranges outside these regions can lead to abnormal generations. These results suggest that, similar to LLMs, MLLMs also contain stable reusable latent regions, while the exact location of such regions is model-dependent. We further conduct this analysis on additional datasets, with results reported in the appendix.

\begin{figure}[h]
    \centering
    \includegraphics[width=\columnwidth]{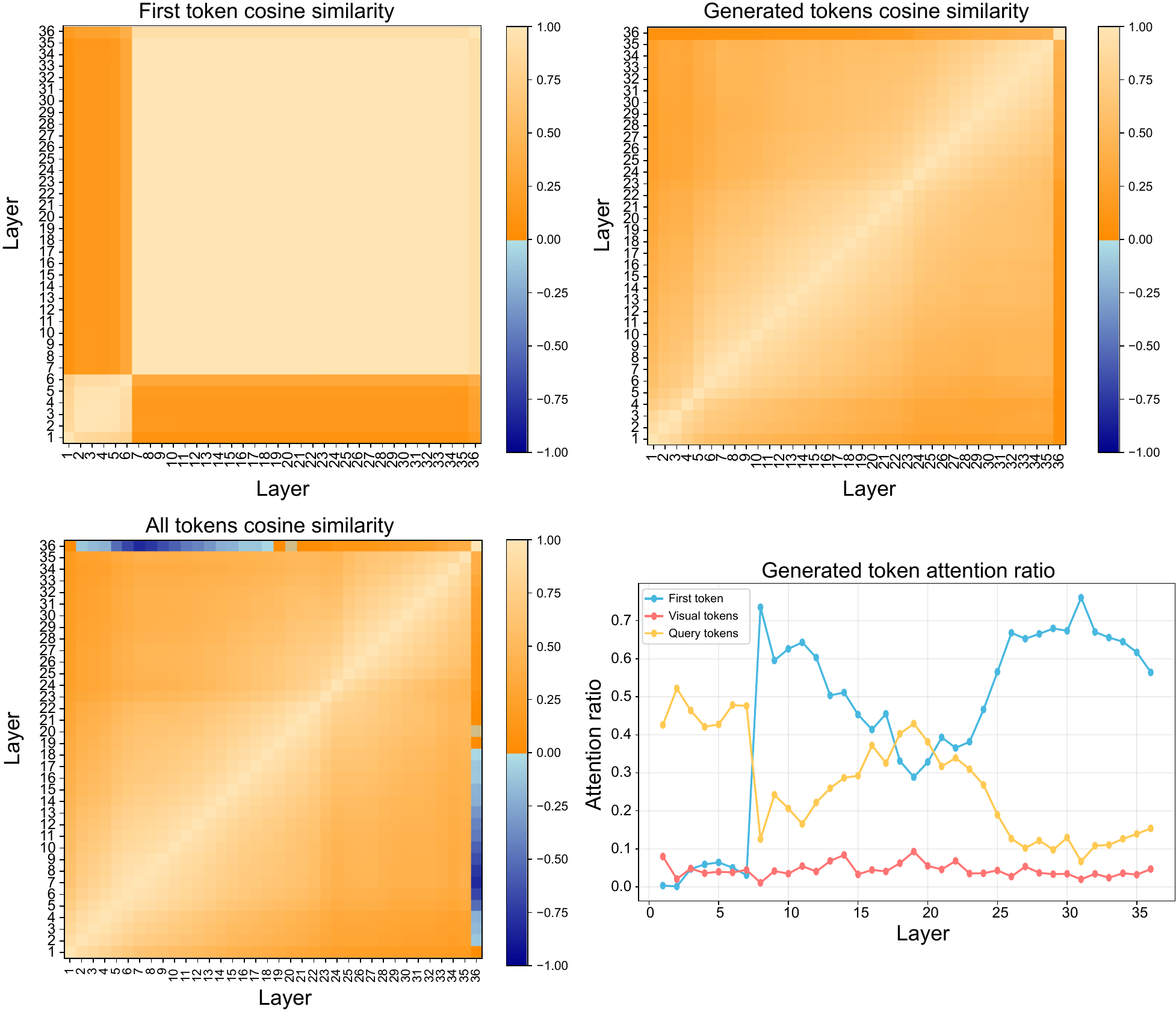}
    \caption{
Layer-wise cosine similarity of hidden states and attention distribution in Qwen3-VL-8B~\cite{bai2025qwen3}. 
}
    \label{fig:hidden_state}
\end{figure}

We next study why these regions can be safely reused. Taking Qwen3-VL-8B as an example, we analyze the layer-wise hidden-state cosine similarity and attention distribution. As shown in Figure~\ref{fig:hidden_state}, the stable reusable region exhibits strong representational consistency across different token types. In particular, we observe a sharp change in the cosine similarity of the first-token hidden state around layer 7, while layer 8 exactly corresponds to the starting point of the reusable region. This transition also coincides with the first emergence of a strong attention-sink pattern, where generated tokens begin to allocate a large fraction of attention to the first token. Prior studies have shown that attention sinks can serve as global aggregation anchors and indicate a qualitative change in the internal state of transformer models~\cite{queipo2025attention,su2026attention}. Therefore, the first appearance of the attention sink provides a natural signal for identifying the beginning of the stable reusable region. The endpoint of the reusable region is also consistent across models: it is always the penultimate layer. This is reasonable because the final layer is more directly coupled with the output prediction head and tends to convert the refined latent representation into logits. As a result, revisiting the final layer can more easily perturb the output-oriented representation and disrupt normal generation. Based on these observations, we identify the stable reusable region as the interval between the first layer where the attention sink emerges and the penultimate layer. Similar phenomena are observed in the other models, with detailed results provided in the appendix.

We further examine whether re-forwarding within the stable reusable region significantly changes the original hidden state. For Qwen3-VL-8B, we take the hidden state at the endpoint of the reusable region, i.e., the penultimate layer, and randomly feed it back into previous reusable-layer windows within layers $[8,35]$, where each window has a length of 9. The results are summarized in Table~\ref{tab:reforward_similarity}. The resulting hidden state remains highly similar to the original one, with an average cosine similarity of 0.996. In contrast, feeding the same hidden state back into layer 7 leads to a much larger change, with the similarity dropping to 0.671. Re-forwarding it through early-layer windows within layers $[1,6]$ causes a stronger perturbation, reducing the average similarity to 0.277. 

\begin{table}[h]
\centering
\begin{tabular}{l c}
\toprule
Revisit region & Cosine similarity \\
\midrule
Reusable region $[8,35]$& 0.996 \\
Layer 7 & 0.671 \\
Early-layer region $[1,6]$ & 0.277 \\
\bottomrule
\end{tabular}
\caption{Cosine similarity between the original hidden state and the hidden state after re-forwarding through different regions in Qwen3-VL-8B~\cite{bai2025qwen3}.}
\label{tab:reforward_similarity}
\end{table}

These results suggest that different depth regions of MLLMs play distinct functional roles. The middle-to-late reusable region primarily performs progressive refinement over an already formed multimodal representation, which is consistent with prior findings on the functional organization of transformer layers~\cite{sun2024transformer,wu2025semantic}. Re-forwarding through this region does not substantially alter the semantic state; instead, it provides additional computation to further refine the hidden representation. This property makes the stable reusable region a suitable latent space for repeated inference and motivates our method to let the model look back within this region.

\section{Method}

Based on the observations above, we propose \textbf{Lo}ok \textbf{Ba}ck in the \textbf{L}atent \textbf{S}pace (\textbf{Lobals}), a training-free decoding method that revisits the most visually grounded part of the stable reusable latent band during generation. For an MLLM with $L$ decoder layers, we first identify a reusable band $\mathcal{R}=[s,e]$ for each model, where $s$ is the first layer in which the attention-sink pattern emerges and $e$ is the penultimate layer.

At each decoding step $t$, Lobals performs a normal forward pass and measures how strongly each layer in $\mathcal{R}$ attends to the visual tokens. The visual-attention ratio at layer $\ell$ is computed as
\begin{equation}
\alpha^{(\ell)}_t
=
\frac{1}{H}
\sum_{h=1}^{H}
\sum_{j\in\mathcal{V}}
A^{(\ell)}_{t,h,j},
\quad \ell \in \mathcal{R},
\end{equation}
where $\mathcal{V}$ is the set of visual-token positions, $A^{(\ell)}_{t,h,j}$ is the attention weight from the current token at decoding step $t$ to position $j$ in head $h$ of layer $\ell$, and $H$ is the number of attention heads. This score measures the fraction of attention allocated to visual evidence at each reusable layer.

Given a fixed window length $k$, Lobals evaluates every contiguous window within the reusable band and selects the one with the largest accumulated visual-attention ratio:
\begin{equation}
W_t^\star
=
\arg\max_{W\subseteq \mathcal{R}, |W|=k}
\sum_{\ell\in W}
\alpha^{(\ell)}_t .
\end{equation}
The selected window $W_t^\star=[a_t,b_t]$ corresponds to the local computation region that is most strongly grounded in the image at decoding step $t$, where $b_t=a_t+k-1$.

After identifying the visually grounded window, Lobals performs a latent-space revisit. Specifically, the hidden state obtained after the endpoint of the reusable band in the normal forward pass, $\mathbf{h}^{(e)}_t$, is fed back into the starting layer $a_t$ of the selected window $W_t^\star=[a_t,b_t]$. The model then re-executes the layers from $a_t$ to $b_t$, producing a refined hidden state. This refined hidden state is subsequently passed to the final layer after $e$. Thus, instead of directly following the original single-pass computation path
\begin{equation}
\mathbf{h}^{(s-1)}_t
\rightarrow
\mathbf{h}^{(s)}_t
\rightarrow
\cdots
\rightarrow
\mathbf{h}^{(e)}_t
\rightarrow
\mathbf{h}^{(L)}_t,
\end{equation}
Lobals inserts a look-back latent computation within the model's normal forward pass:
\begin{equation}
\mathbf{h}^{(s)}_t
\rightarrow
\cdots
\rightarrow
\mathbf{h}^{(e)}_t
\rightarrow
W_t^\star
\rightarrow
\mathbf{h}^{(L)}_t.
\end{equation}

This design enables the model to spend additional computation on the latent region that is most attentive to visual evidence before producing the final output-oriented representation. Since the selected window lies inside the stable reusable region, the revisit operation refines the multimodal representation without disrupting the semantic state, thereby improving visual perception in a training-free manner.

\section{Experiments}

\subsection{Experimental Setup}

\textbf{Benchmarks.}
We evaluate Lobals on six multimodal benchmarks covering complementary aspects of vision-language understanding. MME~\cite{fu2023mme}, MMStar~\cite{chen2024we}, and RealWorldQA~\cite{RealworldQA_2024} are used to evaluate general multimodal understanding. POPE, based on MS-COCO~\cite{li2023evaluating}, and MMHal-Bench~\cite{sun2024aligning} are used to evaluate visual hallucination. MMVP~\cite{tong2024eyes} is used to evaluate vision-centric understanding.

\textbf{Models.}
We conduct experiments on three state-of-the-art open-weight MLLMs with different scales and architectures: Qwen3-VL-2B, Qwen3-VL-8B~\cite{bai2025qwen3}, and InternVL3.5-14B~\cite{wang2025internvl3}. We compare Lobals with three representative training-free decoding methods: Visual Contrastive Decoding (VCD)~\cite{leng2024mitigating}, Pay Attention
to Image (PAI)~\cite{liu2024paying}, and Visual Amplification Fusion (VAF)~\cite{yin2025clearsight}.

\textbf{Implementation details.}
Unless otherwise specified, we set the window length to $k=3$ for Qwen3-VL-2B, $k=9$ for Qwen3-VL-8B, and $k=9$ for InternVL3.5-14B. During inference, we use deterministic decoding with a temperature of 0 and set top-$p$ to 1.0. We use LMMS-EVAL~\cite{zhang2025lmms} for evaluation, and all experiments are run on NVIDIA H200 GPUs. More details about implementation and runtime overhead are provided in the appendix.

\subsection{Main Results}

\begin{table*}[t]
  \centering
  \resizebox{\textwidth}{!}{%
  \begin{tabular}{cccccccccc}
    \toprule
    \multirow{2}{*}{Model} & \multirow{2}{*}{Method}
      & RealWorldQA & MMStar & MME & POPE
      & \multicolumn{2}{c}{MMHal}
      & MMVP & Average \\
    \cmidrule(lr){3-3}
    \cmidrule(lr){4-4}
    \cmidrule(lr){5-5}
    \cmidrule(lr){6-6}
    \cmidrule(lr){7-8}
    \cmidrule(lr){9-9}
    \cmidrule(lr){10-10}
    & & Acc. $\uparrow$ & Avg. $\uparrow$ & Score $\uparrow$
      & Acc. $\uparrow$
      & Score $\uparrow$ & Hall. Rate $\downarrow$
      & Acc. $\uparrow$ & $\uparrow$ \\
    \midrule

    \multirow{5}{*}{Qwen3-VL-2B}
    & Base
      & \second{63.92} & 53.46 & 2019.80 & 89.96
      & 3.58 & 44.21\% & 71.67 & 68.47 \\
    & VCD
      & \best{64.71} & \second{54.73} & \best{2044.34} & 89.96
      & \second{3.90} & \second{37.50\%} & 74.33 & \best{70.29} \\
    & PAI
      & 60.00 & 53.43 & 1990.46 & \best{90.31}
      & 3.68 & 45.83\% & \second{75.00} & 68.53 \\
    & VAF
      & 63.66 & \best{54.95} & 2009.18 & 89.04
      & 3.48 & 43.75\% & \best{78.33} & \second{69.29} \\
    & \textbf{Lobals}
      & \best{64.71} & 54.64 & \second{2032.10} & \second{89.97}
      & \best{3.93} & \best{35.53\%} & 74.33 & \best{70.29} \\

    \midrule

    \multirow{5}{*}{Qwen3-VL-8B}
    & Base
      & 69.02 & 63.59 & 2368.80 & 88.90
      & 4.53 & 30.53\% & 79.00 & 76.77 \\
    & VCD
      & \second{69.93} & \second{64.57} & \second{2378.90} & 89.11
      & \second{4.64} & \second{27.08\%} & \second{82.33}
      & \second{78.04} \\
    & PAI
      & 68.63 & \best{64.68} & 2323.12 & 89.02
      & 4.30 & 35.11\% & \best{84.00} & 76.83 \\
    & VAF
      & \best{70.20} & 63.18 & 2354.72 & \best{89.52}
      & \best{4.67} & \best{26.04\%} & 81.67 & 77.75 \\
    & \textbf{Lobals}
      & \second{69.93} & 64.22 & \best{2409.30} & \second{89.26}
      & 4.62 & 29.47\% & \second{82.33} & \best{78.13} \\

    \midrule

    \multirow{5}{*}{InternVL3.5-14B}
    & Base
      & 69.02 & 67.03 & \second{2379.20} & 85.88
      & 4.25 & \second{23.96\%} & \second{80.67} & 76.40 \\
    & VCD
      & 69.02 & 67.04 & 2379.18 & 85.88
      & 4.27 & \best{21.88\%} & \second{80.67} & \second{76.46} \\
    & PAI
      & \second{69.28} & \best{67.71} & 2368.89 & \best{86.67}
      & 3.82 & 36.46\% & \best{81.67} & 75.60 \\
    & VAF
      & \best{69.93} & 66.96 & 2315.82 & 85.34
      & \second{4.29} & \best{21.88\%} & 79.67 & 76.02 \\
    & \textbf{Lobals}
      & \best{69.93} & \second{67.09} & \best{2389.40} & \second{85.99}
      & \best{4.30} & \best{21.88\%} & \second{80.67} & \best{76.78} \\

    \bottomrule
  \end{tabular}
  }
\caption{Main results of Lobals. MME is normalized by 2800 and the MMHal-Bench score by 6 when computing the average score. The best results are highlighted in \best{bold}, and the second-best are \second{underlined}.}
\label{tab:main_general}
\end{table*}

Table~\ref{tab:main_general} compares Lobals with the base models and three representative training-free decoding methods. Lobals achieves the best or tied-best average performance on all three backbones, demonstrating strong overall competitiveness. More importantly, Lobals improves upon or matches the corresponding base model on every reported metric, whereas each competing method causes performance degradation in at least one setting. These results indicate that Lobals achieves competitive and comparatively stable improvements across models and benchmarks.

For general multimodal understanding, Lobals achieves consistent improvements over the base models on RealWorldQA, MMStar, and MME. The advantage is particularly clear on MME, where Lobals achieves the best performance on Qwen3-VL-8B and InternVL3.5-14B. Lobals also improves vision-centric perception and hallucination mitigation. It improves MMVP performance on both Qwen3-VL models and improves POPE accuracy across all three backbones. On MMHal-Bench, Lobals consistently increases the score while reducing the hallucination rate relative to the base model, achieving the best results on Qwen3-VL-2B and InternVL3.5-14B. Overall, Lobals provides competitive performance across diverse benchmarks while avoiding the degeneration observed in the compared training-free decoding methods.
\subsection{Ablation Studies}

\begin{table}[h]
\centering
\normalsize
\setlength{\tabcolsep}{14pt}
\begin{tabular}{ccc}
\toprule
Window Selection & Acc. $\uparrow$ & F1 $\uparrow$ \\
\midrule
\textit{Start}  & 89.08 & 88.44 \\
\textit{Middle} & 89.01 & 88.45 \\
\textit{End}    & 88.26 & 87.32 \\
\textit{Random} & 89.17 & 88.63 \\
Visually grounded & \textbf{89.26} & \textbf{88.75} \\
\bottomrule
\end{tabular}
\caption{Ablation on revisit window selection for Qwen3-VL-8B on POPE. \textit{Random} results are averaged over three runs.}
\label{tab:ablation_window}
\end{table}

\begin{figure*}[!t]
    \centering
    \includegraphics[width=\textwidth]{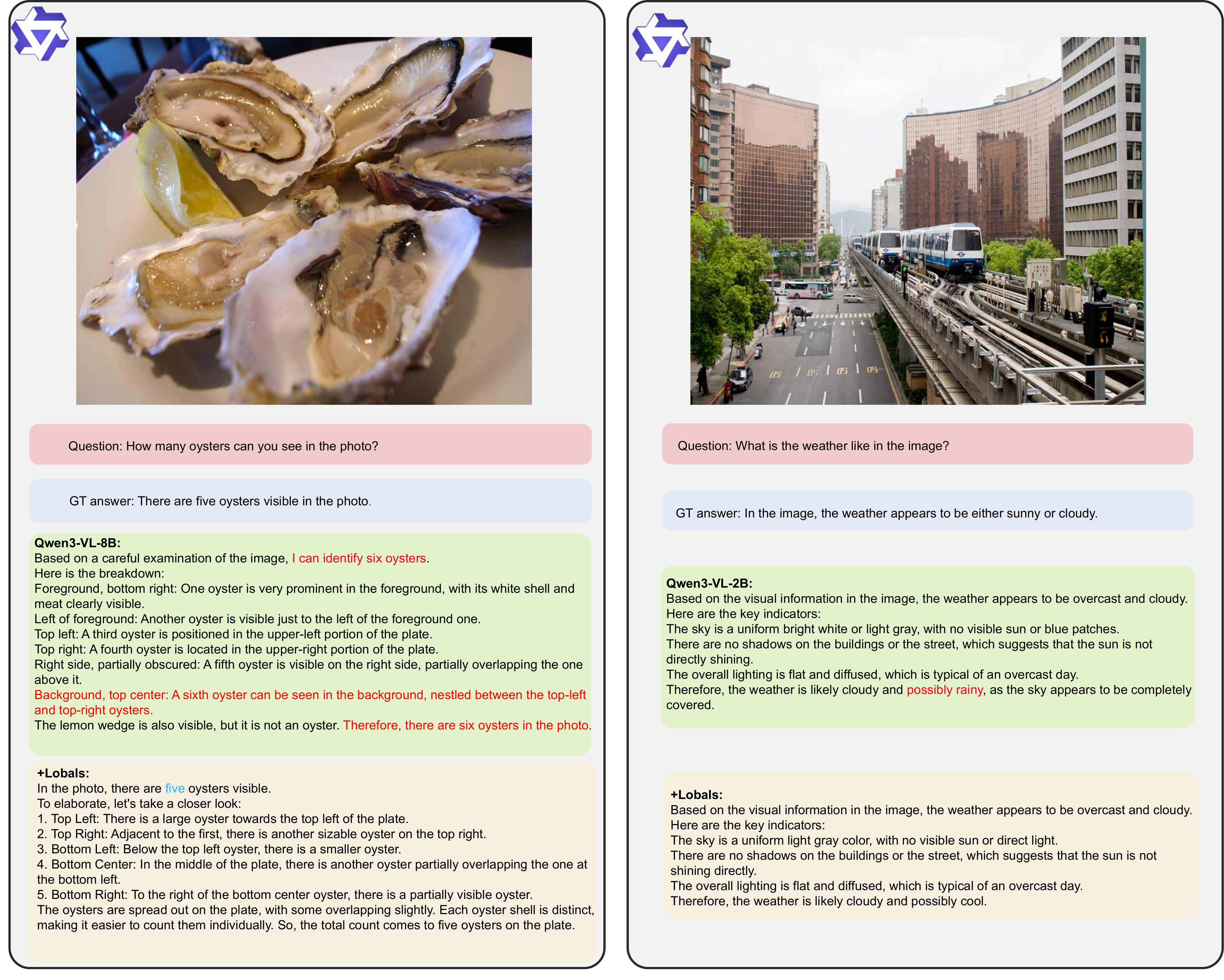}
    \caption{Case studies of Lobals on Qwen3-VL.}
    \label{fig:case}
\end{figure*}

We conduct ablation studies to analyze whether the improvement of Lobals comes from simply adding extra computation within the reusable region, or from adaptively revisiting the visually grounded latent window. We compare Lobals with four baselines. The \textit{Start} strategy always selects the first feasible window in the stable reusable region. The \textit{End} strategy selects the last feasible window whose endpoint is aligned with the endpoint of the reusable region. The \textit{Middle} strategy selects the window located near the center of the feasible window range. The \textit{Random} strategy randomly samples a feasible window from the reusable region, and we report the average result over three runs. Unlike these fixed or random strategies, the \textit{Visually grounded} strategy used by Lobals dynamically selects the contiguous window with the largest accumulated visual-token attention ratio at each decoding step.

\begin{table}[t]
\centering
\normalsize
\setlength{\tabcolsep}{5pt}
\resizebox{\columnwidth}{!}{%
\begin{tabular}{ccccc}
\toprule
Model & Method
& RealWorldQA $\uparrow$
& POPE $\uparrow$
& MME $\uparrow$ \\
\midrule

\multirow{3}{*}{Qwen3-VL-8B}
& Base
& 69.02 & 88.90 & 2368.80 \\
& \textit{Random}
& 69.89 & 89.17 & \best{2416.61} \\
& Lobals
& \best{69.93} & \best{89.26} & 2409.30 \\
\midrule

\multirow{3}{*}{Qwen3-VL-2B}
& Base
& 63.92 & 89.96 & 2019.80 \\
& \textit{Random}
& \textcolor{gray}{63.53} & \textcolor{gray}{89.85} & 2021.84 \\
& Lobals
& \best{64.71} & \best{89.97} & \best{2032.10} \\
\midrule

\multirow{3}{*}{InternVL3.5-14B}
& Base
& 69.02 & 85.88 & 2379.20 \\
& \textit{Random}
& 69.37 & 85.94 & 2385.38 \\
& Lobals
& \best{69.93} & \best{85.99} & \best{2389.40} \\
\bottomrule
\end{tabular}
}
\caption{Comparison of random and visually grounded window selection across models and benchmarks.}
\label{tab:ablation_random}
\end{table}

The \textit{Random} strategy performs competitively on POPE with Qwen3-VL-8B, suggesting that multiple windows within the stable reusable region can provide useful additional computation. However, the visually grounded strategy still achieves the best result. To test whether this advantage generalizes, we further compare random and visually grounded selection across models and benchmarks in Table~\ref{tab:ablation_random}. Lobals outperforms random selection in eight of the nine settings and never falls below the corresponding base model, whereas random selection degrades Qwen3-VL-2B on both RealWorldQA and POPE. These results suggest that re-forwarding within the stable reusable region provides the underlying opportunity for improvement, but arbitrarily allocating the additional computation is not always reliable. By dynamically selecting the window with the strongest visual-token attention, Lobals produces more reliable gains across different settings. Appendix~\ref{sec:expanded_ablation} reports an expanded comparison of all four selection baselines across three models and five benchmarks.

\subsection{Case Studies}

Figure~\ref{fig:case} presents two qualitative examples from MMHal-Bench~\cite{sun2024aligning} to illustrate how Lobals affects visual perception. 
In the first example, the question asks the model to count the number of oysters in the image. The baseline Qwen3-VL-8B predicts six oysters and introduces an additional ``background'' oyster between the two upper oysters. In contrast, Lobals correctly answers that there are five oysters and provides a spatially grounded explanation by enumerating the visible oysters according to their positions on the plate. This example shows that Lobals can reduce object-level hallucination in fine-grained counting scenarios, especially when objects are partially overlapping or visually cluttered. 

The second example focuses on weather recognition. Although the baseline Qwen3-VL-2B correctly observes an overcast sky, it further infers that the weather is ``possibly rainy,'' which is not directly supported by the image. Lobals yields a more conservative and visually faithful response: it identifies the sky as cloudy from visible cues such as the gray sky and diffused lighting, while avoiding the unsupported rainy-weather inference. This suggests that Lobals improves correctness and better grounds explanations in the image.

\subsection{Sensitivity to Window Length}
\begin{figure}[h]
    \centering
    \includegraphics[width=\columnwidth]{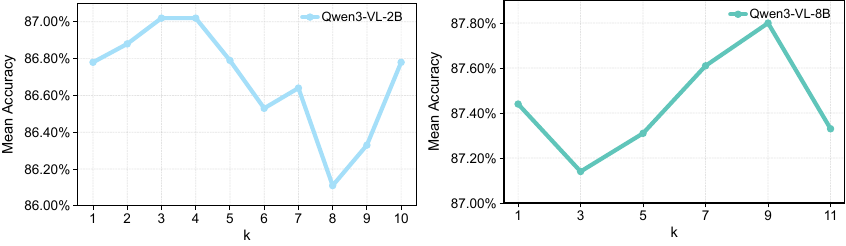}
    \caption{Sensitivity to the revisit window length $k$ on POPE-GQA.}
    \label{fig:k}
\end{figure}
The revisit window length $k$ determines the number of layers revisited at each decoding step, thereby controlling the amount of additional latent computation introduced by Lobals. We evaluate different values of $k$ on POPE-GQA for the models used in our main experiments, with the results for InternVL3.5-14B reported in the appendix.

As shown in Figure~\ref{fig:k}, Lobals is generally stable across a range of window lengths, but the optimal choice of $k$ is model-dependent. Qwen3-VL-2B achieves its best performance with a relatively short window ($k$ = 3), while Qwen3-VL-8B obtains better results with a longer window ($k$ = 9). However, performance does not improve monotonically as $k$ increases. Increasing the window length beyond the optimal value does not lead to better performance.

\subsection{Attention Analysis During Latent Revisit}

To characterize how latent-space revisit is associated with changes in visual attention, we analyze the attention distribution of the last token before, during, and after the revisit process. Specifically, we use Qwen3-VL-8B as a representative model, take the hidden state after layer 35, and feed it back into the selected middle-layer window. We then measure how the last token allocates attention to the first token, visual tokens, and query tokens during the original and revisit passes.

\begin{figure}[h]
    \centering
    \includegraphics[width=\columnwidth]{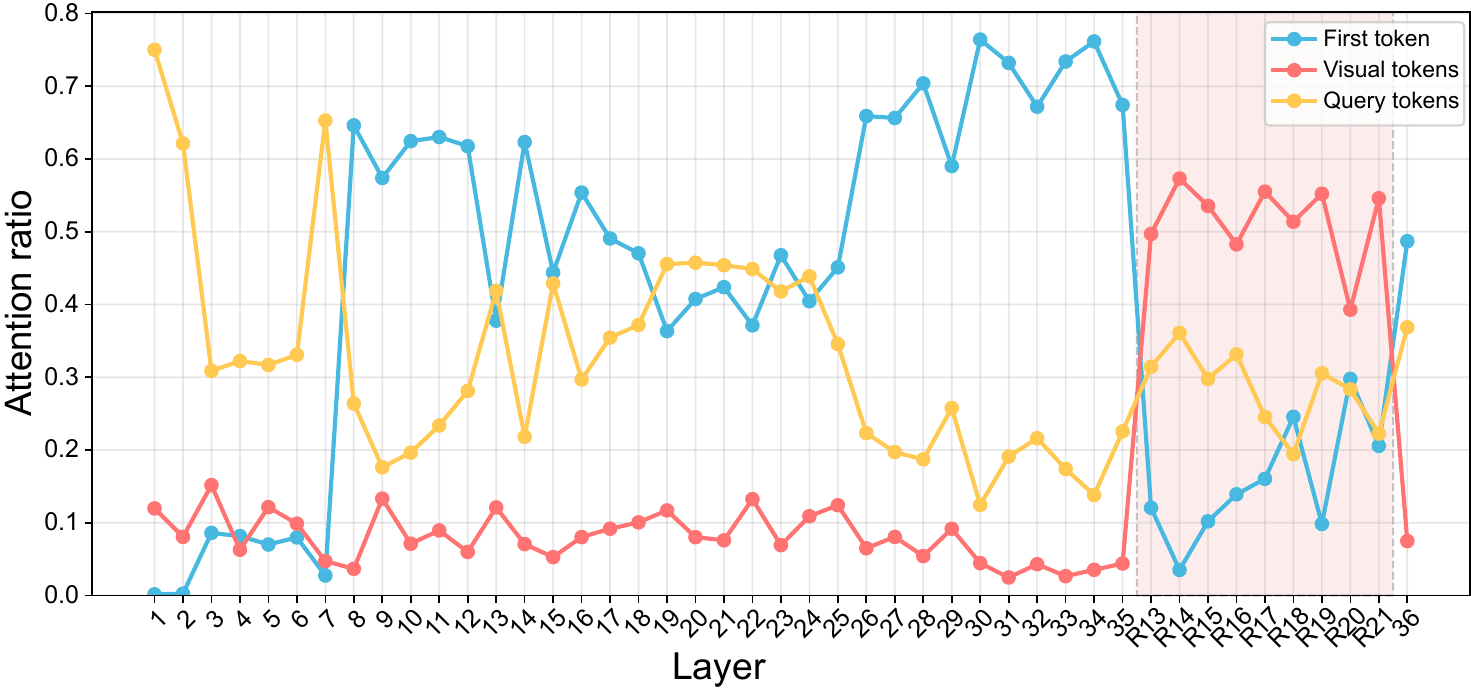}
    \caption{Attention distribution of the last token before, during, and after the revisit process on Qwen3-VL-8B. The shaded region denotes the revisit window.}
    \label{fig:revisit_attention}
\end{figure}

As shown in Figure~\ref{fig:revisit_attention}, the normal forward pass exhibits a clear attention-sink pattern. After the attention sink first emerges around layer 8, a large fraction of attention is consistently assigned to the first token. Meanwhile, attention to visual tokens remains relatively low and further decreases as the model proceeds to deeper layers.

The revisit process substantially changes this allocation pattern. When the hidden state is re-injected into the revisit window, the dominance of the first token is greatly reduced. Across revisit layers R13 to R21, the average first-token attention drops to 15.6\%, whereas the average visual-token attention increases to 51.7\%, making visual tokens the dominant attention target.

This observation provides a plausible explanation for the effectiveness of Lobals. The revisit window reopens a visually interactive latent region, allowing the model to refresh its access to image information before continuing to the final output-oriented layers. This may help strengthen visual perception at a stage where the normal forward pass has already shifted much of its attention away from visual tokens. This is consistent with recent methods for mitigating hallucination, which seek to improve visual faithfulness by strengthening the model's attention to visual tokens~\cite{liu2024paying,yin2025clearsight,yang2026look}. Lobals achieves a related effect through latent-space re-forwarding, without directly modifying attention weights. It should be noted that visual-token attention is not a causal measure of visual grounding; we therefore interpret this analysis only as correlational evidence consistent with the proposed mechanism.

\section{Conclusion}
In this work, we investigate whether MLLMs can revisit previous layers during inference and benefit from such additional latent computation. We show that MLLMs contain stable reusable regions extending from the first layer exhibiting the attention-sink pattern to the penultimate layer, where re-forwarding preserves normal generation. Based on this observation, we propose Lobals, a training-free decoding method that dynamically selects the most visually grounded window and re-forwards through it to strengthen visual evidence before final prediction. Experiments across diverse multimodal settings show that Lobals can improve visual perception without modifying model parameters or introducing extra tools.

\section*{Limitations}

Although Lobals improves visual perception in a training-free manner, it has three limitations. 
First, our current method selects the revisit window based on the visual-attention ratio. 
While this provides a simple and effective heuristic for identifying visually grounded layers, attention may not fully reflect how much the model actually relies on visual tokens during generation~\cite{wen2025stop,zhang2025redundancy}. Therefore, the selected window may not always be optimal. 

Second, we identify the stable reusable region based on the first layer where the attention-sink pattern emerges. 
This criterion works well for the models studied in this paper, but it may not generalize to all future MLLMs. 
Recent model designs may mitigate or weaken attention-sink behaviors, making this signal less reliable~\cite{qiu2026gated,qwen3technicalreport}. 
For such models, a new model-agnostic method is needed to determine the stable reusable region.

Third, the revisit window length $k$ is a model-specific hyperparameter that requires selection. Automatically determining or dynamically adapting $k$ during inference is an important direction for future work.

\bibliography{custom}

\newpage
\appendix

\section{Detailed Implementation}
\textbf{KV Cache.}
Lobals uses the standard autoregressive KV cache during decoding.
At each decoding step, we first perform a normal forward pass to obtain the KV cache for the current token. The subsequent revisit computation only refines the hidden state used for next-token prediction and does not write any new keys or values into the persistent KV cache. Therefore, the KV cache used in the next decoding step is still the one produced by the normal forward pass, ensuring consistency with standard cached decoding.

\textbf{Stable reusable region.}
The starting layer of the stable reusable region is defined as the first layer where the attention-sink pattern emerges, and the endpoint is set to the penultimate layer. Following this criterion, the stable reusable regions are $[8,35]$ for Qwen3-VL-8B, $[4,27]$ for Qwen3-VL-2B, and $[8,39]$ for InternVL3.5-14B. 

\textbf{Evaluation.} 
We evaluate all benchmarks with LMMS-EVAL, except for MMHal-Bench.
For MMHal-Bench, we follow the official evaluation setting, but use GPT-5.5 as the LLM judge. Unlike text-only judging, we also provide the corresponding image to GPT-5.5 during evaluation, allowing the judge to assess whether the model response is supported by the visual evidence.

\section{Runtime Overhead}

Lobals introduces controllable overhead during decoding. 
At each decoding step, the normal forward pass is still executed once, and the visual-attention ratios used for window selection are obtained from this pass. 
Thus, window selection only requires a lightweight sliding-window search and does not introduce another full forward pass.

The main extra cost comes from re-forwarding through the selected window. 
If the model has $L$ decoder layers and the revisit window length is $k$, Lobals adds $k+1$ extra layer evaluations per generated token, since the reusable region ends at the penultimate layer in our setting. 
The layer-level overhead is therefore approximately $(k+1)/L$. The runtime cost scales linearly with the window length $k$ and the number of generated tokens.

For example, Qwen3-VL-8B has 36 decoder layers, and we use $k=9$ in our main experiments. In this case, Lobals performs about $(36+10)/36 \approx 1.28$ times as much computation as standard decoding.

We further measure the real end-to-end runtime of Qwen3-VL-8B on MMHal-Bench using an NVIDIA H200. As shown in Table~\ref{tab:runtime}, Lobals has 1.70× the baseline wall-clock latency while retaining 63.8\% of the baseline throughput.

\begin{table}[h]
\centering
\small
\begin{tabular}{lcc}
\toprule
Method
& Latency / Sample (s) $\downarrow$
& Output Tokens/s $\uparrow$ \\
\midrule
Base   & 2.53 & 40.08 \\
Lobals & 4.30 & 25.56 \\
\bottomrule
\end{tabular}
\caption{End-to-end runtime of Qwen3-VL-8B on MMHal-Bench using an NVIDIA H200.}
\label{tab:runtime}
\end{table}

\section{Benchmark Details}

\textbf{MME}~\cite{fu2023mme} is a comprehensive benchmark for evaluating multimodal perception and cognition.
It contains a set of yes-or-no questions covering different visual abilities, such as object existence, color, counting, position, scene understanding, and commonsense reasoning. Following the standard protocol, we report the overall MME score, where a higher score indicates better multimodal understanding.

\textbf{MMStar}~\cite{chen2024we} is a multiple-choice benchmark designed to evaluate general multimodal understanding. It covers diverse capability dimensions, including visual coarse perception, visual fine-grained perception, and visual mathematical reasoning. We report accuracy as the evaluation metric.

\textbf{RealWorldQA}~\cite{RealworldQA_2024} evaluates visual question answering in real-world scenarios. The benchmark focuses on practical image understanding, including object recognition, spatial relations, and scene-level reasoning. We report accuracy following the official evaluation setting.

\textbf{POPE}~\cite{li2023evaluating} is used to evaluate object hallucination in MLLMs. It formulates hallucination detection as a binary yes-or-no question answering task. We evaluate on the random, popular, and adversarial splits from the MS-COCO source and report the averaged accuracy and F1 score.

\textbf{MMHal-Bench}~\cite{sun2024aligning} evaluates hallucination in open-ended multimodal responses.
It contains image-question pairs that require the model to generate free-form answers, and the responses are judged according to both correctness and visual informativeness. Following the official protocol, we report the average score and hallucination rate. 

\begin{figure*}[!t]
    \centering
    \includegraphics[width=\textwidth]{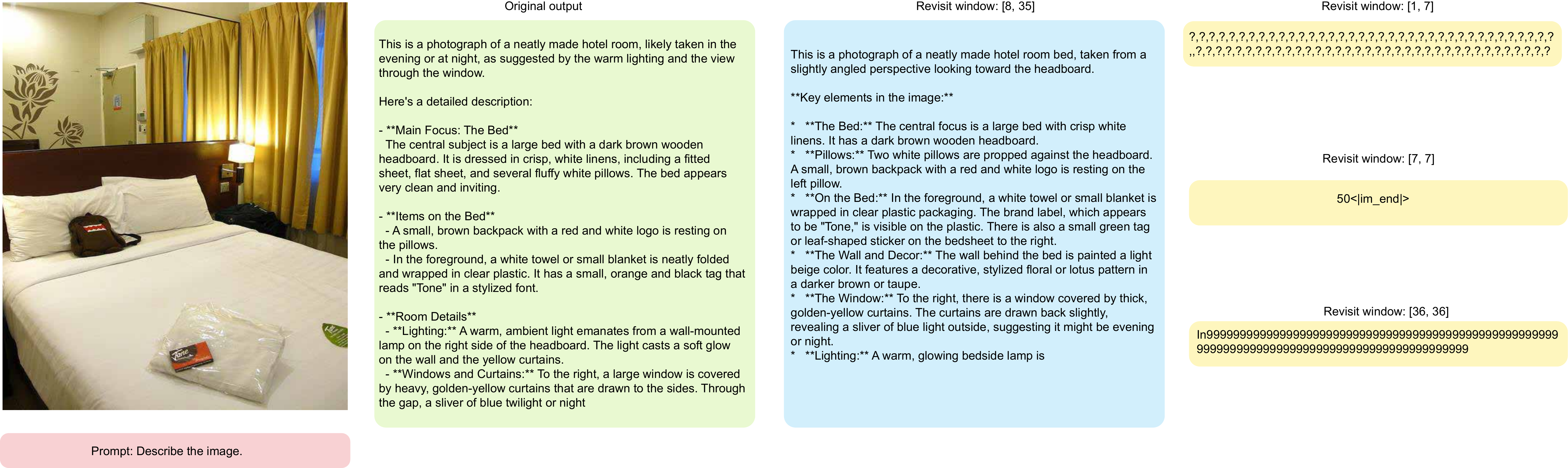}
    \caption{Case study of fluency judgment on Qwen3-VL-8B under different revisit windows.}
    \label{fig:case-fluency}
\end{figure*}

\textbf{MMVP}~\cite{tong2024eyes} evaluates vision-centric understanding in MLLMs.
It contains visually challenging VQA examples built from CLIP-blind image pairs, where images are close in the CLIP embedding space but differ in important visual details. These examples require models to rely on fine-grained visual perception. We report accuracy as the evaluation metric.

\section{Fluency Judgment Details}
\textbf{Fluency judgment prompt}. To investigate whether MLLMs contain latent
regions that can be revisited without disrupting
normal generation, we use GPT-5.1 to judge the fluency of the generated response after latent-space revisit. 
The prompt template is shown below.
\begin{center}
\setlength{\fboxsep}{9pt}
\setlength{\fboxrule}{0.8pt}
\fbox{%
\begin{minipage}{\dimexpr\linewidth-2\fboxsep-2\fboxrule\relax}
\small
You are a strict fluency judge.

Your task is to decide whether the given model output is fluent.

Evaluate only fluency: naturalness, readability, grammaticality, coherence, and whether the text is understandable.

Do NOT evaluate:
\begin{itemize}
    \item factual correctness
    \item completeness
    \item helpfulness
    \item whether the answer fully satisfies the user
    \item whether the final sentence is unfinished due to max token truncation
\end{itemize}

Important rule:

If the output appears truncated because of length limits, do NOT mark it as non-fluent solely for being cut off. Judge only the text before the truncation. 

Output format:

Return only one character:

1 = fluent

0 = not fluent

Model output:

\texttt{\{model\_output\}}
\end{minipage}%
}
\end{center}

\textbf{Case study}. Figure~\ref{fig:case-fluency} presents a representative example on Qwen3-VL-8B with a maximum generation length of 256 tokens. Both the original forward pass and the revisit through the stable reusable window $[8,35]$ produce fluent and coherent descriptions. In contrast, revisiting unstable regions causes clear generation collapse, including repeated symbols, digits, and abnormal token fragments.

\begin{figure*}[t]
    \centering
    \includegraphics[width=\textwidth]{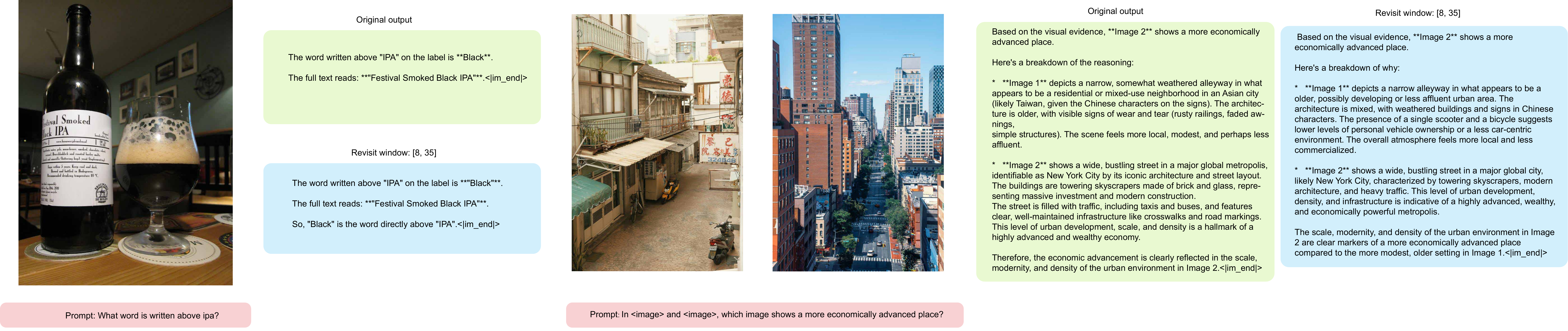}
    \caption{Case studies of fluency judgment on TextVQA and Mantis-Eval with Qwen3-VL-8B.}
    \label{fig:case-more}
\end{figure*}

\textbf{Additional datasets}. The fluency analysis on COCO, together with the results across multiple downstream benchmarks, supports the effectiveness of our rule for identifying the stable reusable region---from the first layer exhibiting the attention-sink pattern to the penultimate layer. To further validate this rule under different task settings, we randomly sampled 50 examples from TextVQA~\cite{textvqa} for OCR and 50 examples from Mantis-Eval~\cite{jiang2024mantisinterleavedmultiimageinstruction} for multi-image tasks. All sampled examples continued to produce normal outputs when re-forwarding was applied within the identified stable reusable region. Figure~\ref{fig:case-more} presents two cases using Qwen3-VL-8B with the revisit window $[8,35]$.

\section{Expanded Ablation Study}
\label{sec:expanded_ablation}

We extend the window-selection ablation to Qwen3-VL-2B, Qwen3-VL-8B, and InternVL3.5-14B on POPE, RealWorldQA, MMVP, MMStar, and MME. Table~\ref{tab:expanded_ablation} compares the base model, random selection, fixed windows at the start, middle, or end of the stable reusable region, and the attention-based selection used by Lobals. These strategies follow the definitions in the main ablation study; random-selection results are averaged over three runs.

\begin{table*}[t]
\centering
\small
\setlength{\tabcolsep}{7pt}
\begin{tabular}{llrrrrr}
\toprule
Model & Method & POPE & RQA & MMVP & MMStar & MME \\
\midrule
\multirow{6}{*}{Qwen3-VL-2B} & Base & 89.96 & 63.92 & 71.67 & 53.46 & 2019.80 \\
 & Random & 89.85$\downarrow$ & 63.53$\downarrow$ & 74.00 & 54.71 & 2021.84 \\
 & Start & 89.81$\downarrow$ & \textbf{64.71} & 74.00 & \textbf{54.74} & 2026.00 \\
 & Middle & 89.83$\downarrow$ & 64.44 & \textbf{74.33} & 54.63 & 2031.98 \\
 & End & 89.38$\downarrow$ & 56.99$\downarrow$ & \textbf{74.33} & 53.72 & 1966.77$\downarrow$ \\
 & Lobals & \textbf{89.97} & \textbf{64.71} & \textbf{74.33} & 54.64 & \textbf{2032.10} \\
\midrule
\multirow{6}{*}{Qwen3-VL-8B} & Base & 88.90 & 69.02 & 79.00 & 63.59 & 2368.80 \\
 & Random & 89.17 & 69.89 & \textbf{82.33} & 64.05 & \textbf{2416.61} \\
 & Start & 89.08 & 69.02 & 82.00 & 64.15 & 2406.86 \\
 & Middle & 89.01 & 69.41 & 82.00 & \textbf{64.45} & 2399.87 \\
 & End & 88.26$\downarrow$ & 69.80 & 82.00 & 63.79 & 2373.22 \\
 & Lobals & \textbf{89.26} & \textbf{69.93} & \textbf{82.33} & 64.22 & 2409.30 \\
\midrule
\multirow{6}{*}{InternVL3.5-14B} & Base & 85.88 & 69.02 & \textbf{80.67} & 67.03 & 2379.20 \\
 & Random & 85.94 & 69.37 & 80.22$\downarrow$ & 66.88$\downarrow$ & 2385.38 \\
 & Start & 85.82$\downarrow$ & 69.28 & \textbf{80.67} & 66.86$\downarrow$ & 2370.88$\downarrow$ \\
 & Middle & 85.93 & 69.28 & \textbf{80.67} & 67.03 & \textbf{2394.86} \\
 & End & 85.97 & \textbf{69.93} & 80.33$\downarrow$ & 66.80$\downarrow$ & 2388.28 \\
 & Lobals & \textbf{85.99} & \textbf{69.93} & \textbf{80.67} & \textbf{67.09} & 2389.40 \\
\bottomrule
\end{tabular}
\caption{Expanded ablation of revisit-window selection. RQA denotes RealWorldQA. The best result for each model and benchmark is in \textbf{bold}; $\downarrow$ marks a score below the corresponding base model.}
\vspace{-1.2em}
\label{tab:expanded_ablation}
\end{table*}

The expanded results show that multiple windows within the reusable region can improve performance. However, random or fixed selection can also degrade performance: for example, the end window reduces Qwen3-VL-2B's RealWorldQA accuracy from 63.92 to 56.99, while random selection reduces InternVL3.5-14B's MMVP accuracy from 80.67 to 80.22. In contrast, Lobals improves upon or matches the base model in all 15 model--benchmark settings. Although individual alternatives achieve higher scores on some benchmarks, attention-based selection provides more reliable gains across the evaluated settings.

\section{More Results}
Table~\ref{tab:llava_next_results} reports additional results on LLaVA-NeXT-13B~\cite{liu2024llavanext}. Lobals improves performance across all evaluated benchmarks, including a higher MMHal-Bench score and a lower hallucination rate. These results extend our evaluation to another MLLM backbone.

\begin{figure}[h]
    \centering
    \includegraphics[width=\columnwidth]{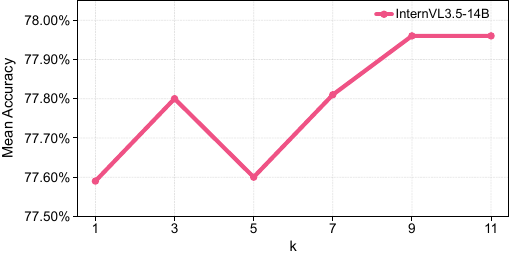}
    \caption{Sensitivity to the revisit window length $k$ on POPE-GQA for InternVL3.5-14B.}
    \label{fig:k-internvl}
\end{figure}

Figure~\ref{fig:k-internvl} supplements the window-length sensitivity analysis with results for InternVL3.5-14B on POPE-GQA, showing how performance varies with the revisit window length $k$.

\begin{figure}[h]
    \centering
    \includegraphics[width=\columnwidth]{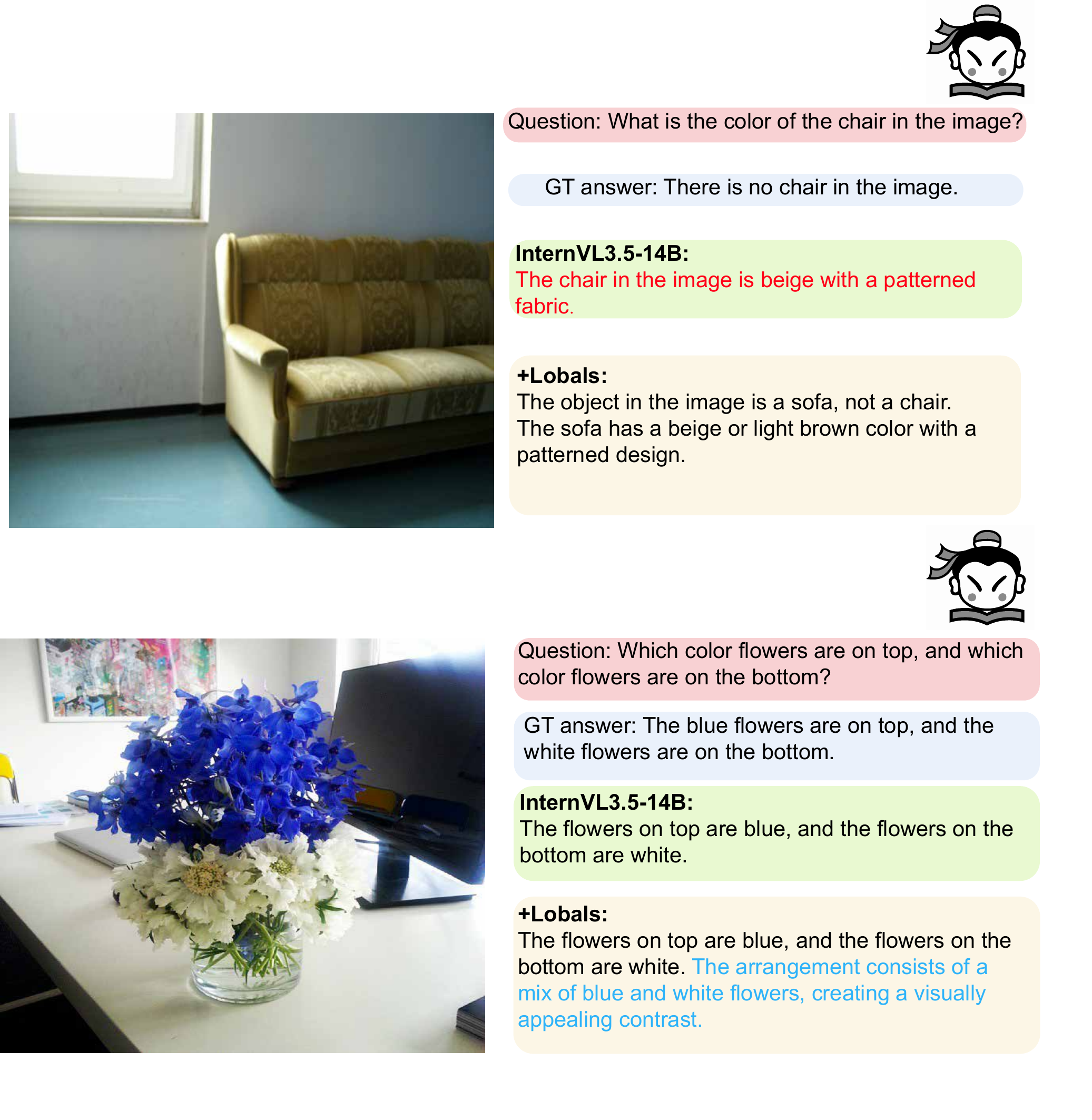}
    \caption{
Case studies of Lobals on InternVL3.5-14B from MMHal-Bench.
Red text highlights hallucinated content in the baseline response, while blue text highlights additional details introduced by Lobals.
}
    \label{fig:case-internvl}
\end{figure}

Figure~\ref{fig:case-internvl} presents qualitative examples on InternVL3.5-14B from MMHal-Bench, illustrating how Lobals reduces hallucinated content and provides additional visually grounded details.

\begin{figure}[h]
    \centering
    \includegraphics[width=\columnwidth]{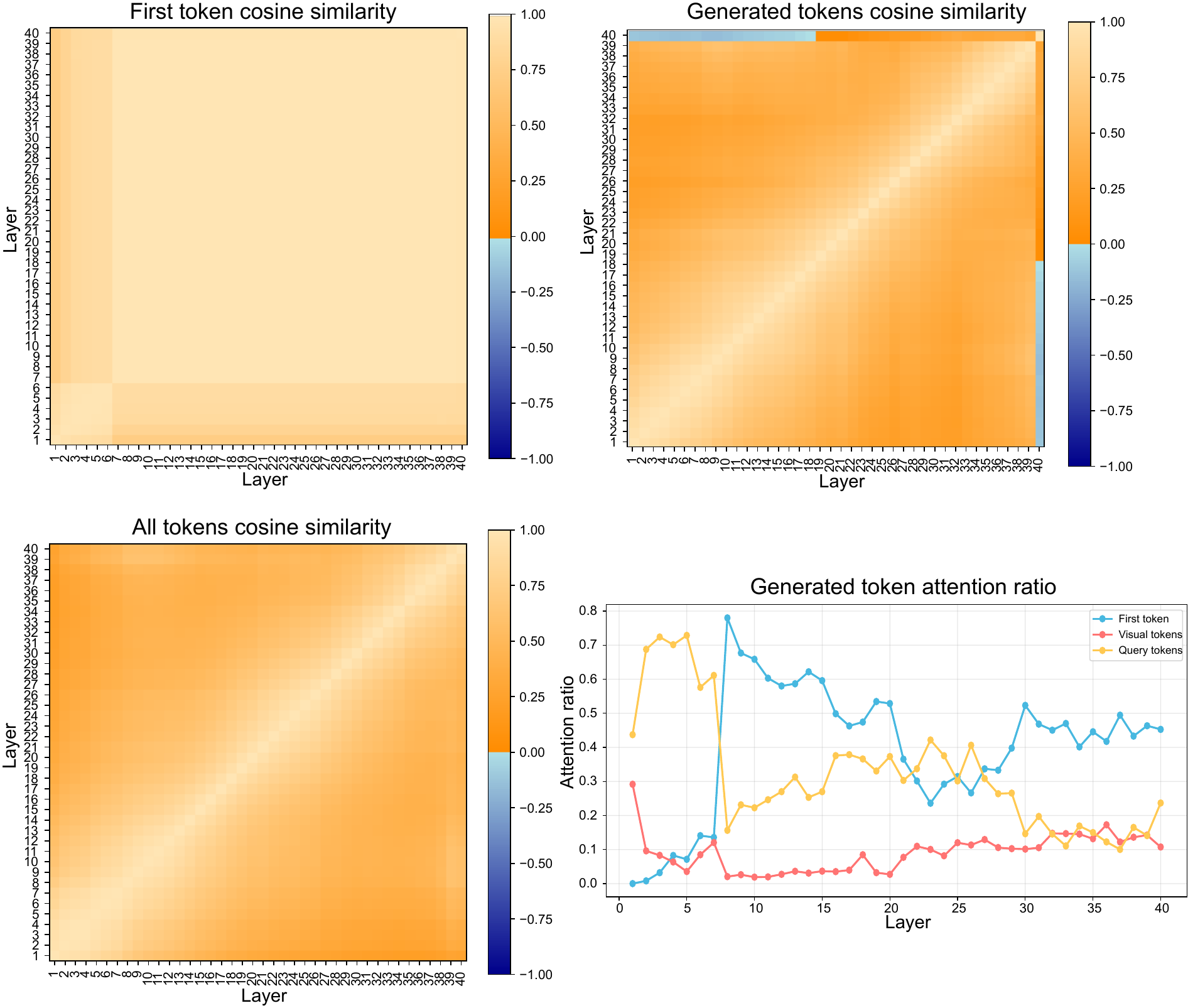}
    \caption{Layer-wise cosine similarity of hidden states and attention distribution in InternVL3.5-14B.}
    \label{fig:state-internvl}
\end{figure}

\begin{figure}[h]
    \centering
    \includegraphics[width=\columnwidth]{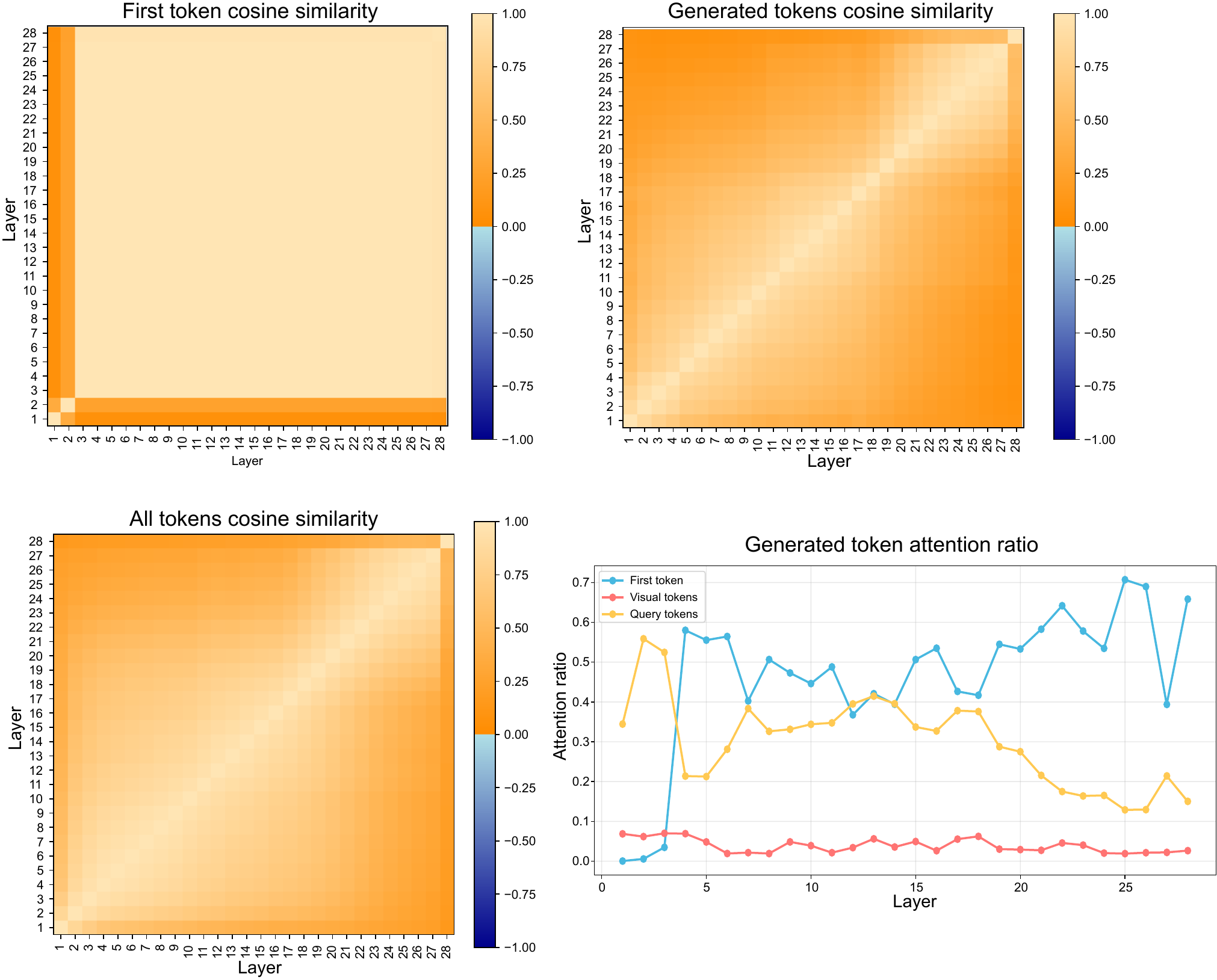}
    \caption{Layer-wise cosine similarity of hidden states and attention distribution in Qwen3-VL-2B.}
    \label{fig:state-qwen2b}
\end{figure}

\begin{figure}[h]
    \centering
    \includegraphics[width=\columnwidth]{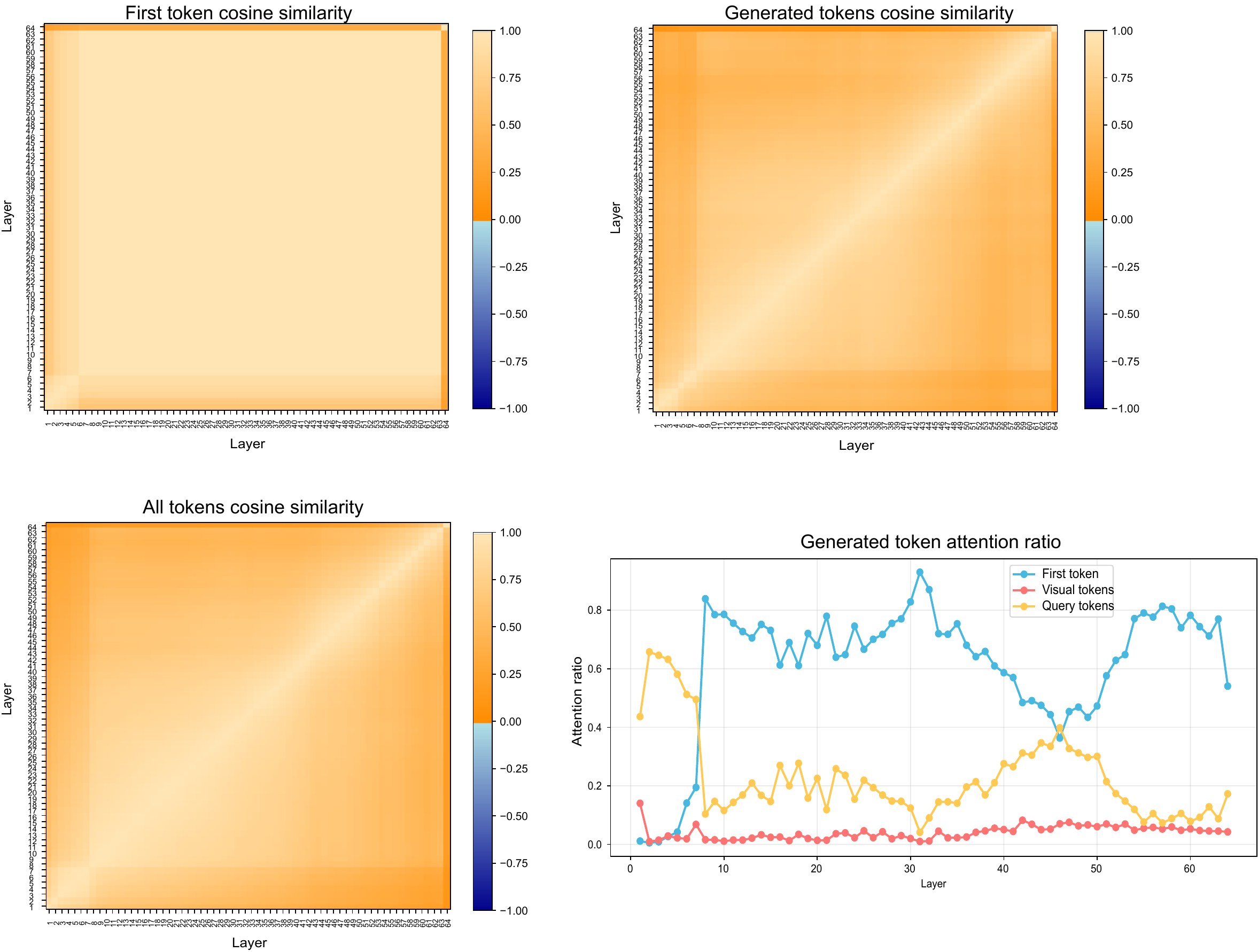}
    \caption{Layer-wise cosine similarity of hidden states and attention distribution in Qwen3-VL-32B.}
    \label{fig:state-qwen32b}
\end{figure}

Figures~\ref{fig:state-internvl}, \ref{fig:state-qwen2b}, and \ref{fig:state-qwen32b} provide additional layer-wise hidden-state cosine similarity and attention analyses for InternVL3.5-14B, Qwen3-VL-2B, and Qwen3-VL-32B, respectively.

\begin{table}[t]
\centering
\normalsize
\begin{tabular}{llcc}
\toprule
\multicolumn{2}{c}{Benchmark} & Base & Lobals \\
\midrule

\multicolumn{2}{l}{MMVP $\uparrow$}
& 66.67
& \textbf{67.00} \\
\midrule

\multicolumn{2}{l}{RealWorldQA $\uparrow$}
& 55.82
& \textbf{57.78} \\
\midrule

\multicolumn{2}{l}{MMStar $\uparrow$}
& 40.53
& \textbf{40.83} \\
\midrule

\multicolumn{2}{l}{POPE $\uparrow$}
& 86.92
& \textbf{88.20} \\
\midrule

\multicolumn{2}{l}{MME $\uparrow$}
& 1801.12
& \textbf{1834.20} \\
\midrule

\multirow{2}{*}{MMHal}
& Score $\uparrow$
& 3.39
& \textbf{3.45} \\
& Hall. Rate $\downarrow$
& 40.62\%
& \textbf{39.58\%} \\

\bottomrule
\end{tabular}
\caption{Results on LLaVA-NeXT-13B. The best results are highlighted in \textbf{bold}.}
\label{tab:llava_next_results}
\end{table}

\end{document}